\documentclass[conference]{IEEEtran}
\ifCLASSINFOpdf
\usepackage[pdftex]{graphicx}
\DeclareGraphicsExtensions{.pdf,.jpeg,.png}
\else
\fi
\usepackage{tabularx,calc}
\usepackage{color}
\usepackage{tikz,amsmath, amssymb,bm,color}
\usepackage{pgfplots}
\usepackage{environ}
\usetikzlibrary{shapes,arrows}
\usetikzlibrary{calc}

\makeatletter
\newsavebox{\measure@tikzpicture}
\NewEnviron{scaletikzpicturetowidth}[1]{%
	\def\tikz@width{#1}%
	\begin{lrbox}{\measure@tikzpicture}%
		\BODY
	\end{lrbox}%
	\pgfmathparse{#1/\wd\measure@tikzpicture}%
	\BODY
}
\makeatother
\usepackage{ifthen}
\newboolean{compileforpublish}
\setboolean{compileforpublish}{true}
\newboolean{isaccepted}
\setboolean{isaccepted}{true}
\ifthenelse{\boolean{isaccepted}}
{%if
\newcommand\copyrighttext{%
	\footnotesize \parbox[t]{.11\textwidth}{\copyright{} \the\year~IEEE.} \parbox[t]{.89\textwidth}{Personal use of this material is permitted. Permission from IEEE must be obtained for all other uses, in any current or future media, including reprinting/republishing this material for advertising or promotional purposes, creating new collective works, for resale or redistribution to servers or lists, or reuse of any copyrighted component of this work in other works.}}
	}
{%else
\newcommand\copyrighttext{%
	\footnotesize \centering This work has been submitted to the IEEE for possible publication.\\ Copyright may be transferred without notice, after which this version may no longer be accessible.}
}

\newcommand\copyrightnotice{%
	\ifthenelse{\boolean{compileforpublish}}
	{
	\begin{tikzpicture}[remember picture,overlay]
	\node[anchor=south,yshift=10.5pt] at (current page.south) {\parbox{\dimexpr\textwidth-\fboxsep-\fboxrule\relax}{\copyrighttext}};
	\end{tikzpicture}%
	}
}

\renewcommand{\IEEEtitletopspaceextra}{20pt}

\begin{document}
\title{Ontology based Scene Creation for the Development of Automated Vehicles}

\author{\IEEEauthorblockN{Gerrit Bagschik, Till Menzel and Markus Maurer}
\IEEEauthorblockA{Institute of Control Engineering\\
Technische Universit\"at Braunschweig\\
Braunschweig, Germany\\
Email: \{bagschik, menzel, maurer\}@ifr.ing.tu-bs.de}
}
\maketitle%
\copyrightnotice%

\begin{abstract}
The introduction of automated vehicles without permanent human supervision demands a functional system description, including functional system boundaries and a comprehensive safety analysis. 
These inputs to the technical development can be identified and analyzed by a scenario-based approach.
Furthermore, to establish an economical test and release process, a large number of scenarios must be identified to obtain meaningful test results. 
Experts are doing well to identify scenarios that are difficult to handle or unlikely to happen. 
However, experts are unlikely to identify all scenarios possible based on the knowledge they have on hand.
Expert knowledge modeled for computer aided processing may help for the purpose of providing a wide range of scenarios.
This contribution reviews ontologies as knowledge-based systems in the field of automated vehicles, and proposes a generation of traffic scenes in natural language as a basis for a scenario creation.
\end{abstract}

\section{Introduction}

Safety assessment of automated driving functions is an emerging topic in the automotive industry.
Several research and development projects show prototypes of automated vehicles in well-defined showcases.
When it comes to series production, the ISO~26262~standard defines a state-of-the-art development process to ensure functional safety.

Automated vehicles will have to fulfill a safe driving task in a high number of operating scenarios.
To comply with the hazard analysis and risk assessment demanded by the ISO~26262~standard, hazardous events ``shall be determined systematically by using adequate techniques'' \cite[Part 3]{ISO_26262_2001}.
Therefore, operating scenarios, in which malfunctioning behavior of the item can be hazardous, have to be identified.
Nowadays, these scenario catalogs are generated by experts during the development process.
Expert-based scenario catalogs can be representative for critical situations.
However, guarantees for completeness in terms of identifying all possible combinations cannot be given.

A main difference from assisted (level 1) or partial automated driving (level 2) to conditional and high automation (level 3 and 4) is the number of scenarios which have to be investigated and defined.
All these scenarios are required to argue that the highly or fully automated driving function is ready for market introduction.
Wachenfeld and Winner~\cite{wachenfeld_release_2016} deduce how many kilometers it takes in field tests for an automated vehicle to be released.
The high number of kilometers may result from the likelihood of encountering possible scenarios which might lead to an accident.
However, they conclude that a scenario-based test process can replace time consuming field tests by shifting test case execution to simulation environments.
Hence, the developers need more experts or more time for the experts to identify all relevant scenarios and combinations.
Expert knowledge is reasonable for investigation of new items in development, but the process of scenario creation by experts is more creative than systematic.
To get traceable and comparable scenarios, the experts need at least a consolidated vocabulary and the same understanding of how scenarios are organized.

Bergenheim et al. \cite{bergenhem_how_2015} further describe a semantic gap in safety validation processes.
Since actual methods for defining requirements do not support the developers in arguing that the developed item is safe, the semantic gap rather occurs already during system development.
The authors claim that the discussion on the \emph{safety of the intended functionality} (SoTIF) can be resolved by a more detailed specification of functional safety requirements or a validation of the functional safety concept towards completeness.
Thus, we propose that requirement specification and validation processes can be supported by systematically identified scenario catalogs in the concept phase of the development process.

Systematically identified scenarios based on expert knowledge can help to improve requirements engineering and safety analysis.
Also, scenarios provide a basis for test case generation for simulation-based testing of automated vehicles.
For this reason, we propose using knowledge-based systems to create scenarios for development of automated driving functions.
Ontologies have successfully been used in the field of automated driving in recent years (cf. Section \ref{sec:related}).
Many of these concepts show, how the use of knowledge-based systems can improve decision making and scene understanding in particular driving scenarios.
Our concept includes knowledge about the design of traffic infrastructure, and the behavior and interaction between traffic participants.
Furthermore, we propose how knowledge about traffic can be represented by a layered concept to divide it into smaller parts and model interactions between the layers.

This contribution is organized as follows:
In the next two parts, the concept of ontologies and related work in the field of automated driving is summarized.
Afterwards, our approach for traffic scene creation is presented and concluded.

\section{Ontologies}

Guarino et al. \cite{staab_what_2009} give a summarized definition of \emph{ontologies} as ``a formal, explicit specification of a shared conceptualization.''
The conceptualization itself is described by Genesereth et al. as ``an abstract, simplified view of the world that we wish to represent for some purpose.
Every knowledge base, knowledge-based system, or knowledge-level agent is committed to some conceptualization, explicitly or implicitly'' \cite{Genesereth31838}.
According to Studer~et~al.~\cite{studer_knowledge_1998}, ontologies can be divided by different levels of generality.
These levels reach from general ontologies, which express domain independent knowledge, to domain and application ontologies, which contain true statements from particular domains like medical applications or electrical engineering or are only applicable to a certain application in a domain.

\begin{figure}[h]
	\begin{center}
	\includegraphics[width=0.47\textwidth]{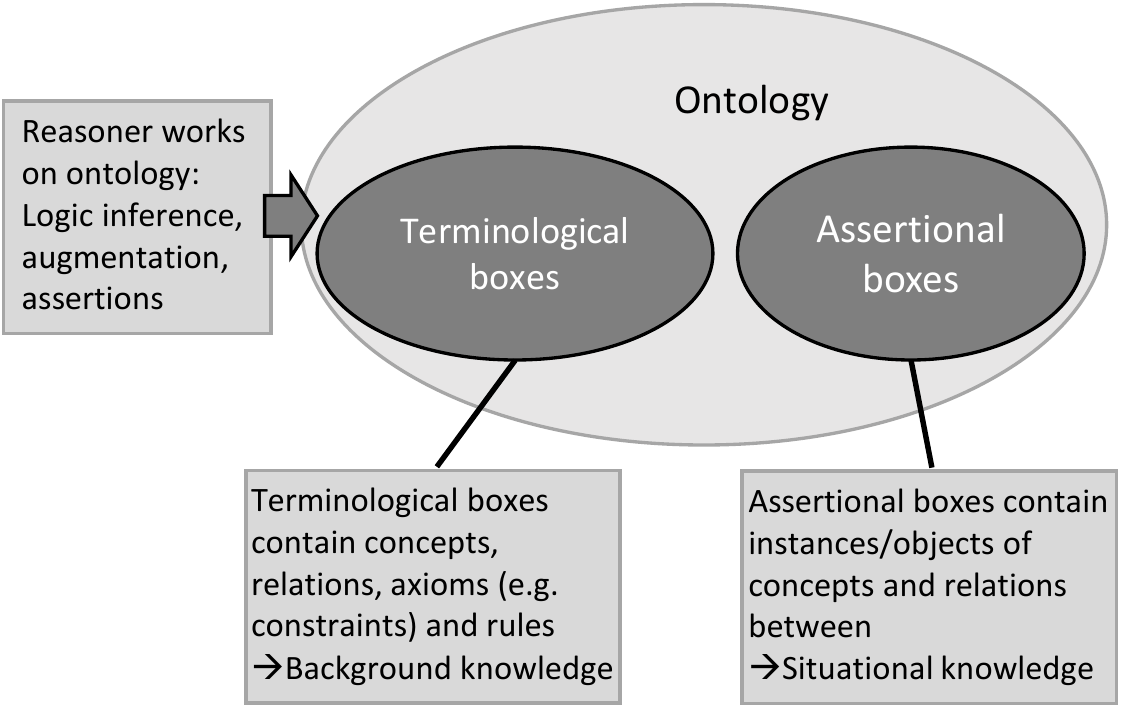}    % The printed column width is 8.4 cm.
	\caption{Architecture of ontologies from H\"ulsen~\cite{hulsen_traffic_2011}} 
	\label{fig:boxes}
	\end{center}
\end{figure}

Fig. \ref{fig:boxes} shows the structure of ontologies with terminological and assertional boxes.
Both types of boxes are combined to form the knowledge base expressed by an ontology.

Terminological boxes describe the concepts of a domain.
These concepts are expressed using hierarchical classes, axioms and properties.
Assertional boxes represent instances of classes and observed facts of situative knowledge about the world.
Following the definitions by Ulbrich~et~al.~\cite{ulbrich_scene_2015}, assertional boxes describe scenes extracted from a given set of possible world entities and relations (defined by terminological boxes).

The concepts stored in an ontology have to be readable and understandable by human experts and computers.
Humans can read the structure of an ontology, including the hierarchy and the axioms, which are expressed by natural language.
For computers, this is accomplished by expressing ontologies in a description logic, which translates every concept of the domain into first order logic. 

A common declarative language (which builds the semantic base for modeling) in the field of ontologies is the \emph{Web Ontology Language} (OWL)\footnote{The abbreviation has switched letters due to a typo on a mailing list and the fact that OWL, like the bird, is easier to remember than WOL.}, which is standardized and maintained by the World Wide Web Consortium (W3C).
It is intentionally designed for applications on the semantic web, where pages can be connected with the meaning of the terms.
For example, this enables better search results based on context and not a pure comparison of terms.
OWL includes three variants of the language, namely OWL Lite, OWL DL (Description Logic), and OWL Full.
The variants differ in the expressivity of the language, which increases from OWL Lite to OWL Full.

Ontologies are stored in first order logic, which allows reasoners to infer knowledge on terminological and assertional boxes.
Reasoners are able to identify hierarchical missing concepts on terminological boxes, check for conflicts in the modeled concepts and check if similar concepts exist over the knowledge base.
Assertional boxes can be checked for consistency and associated with the concepts from terminological boxes to infer situational knowledge, for example from sensor data.
An overview of reasoners and a comparison study is given by Abburu \cite{abburu_survey_2012}.

\section{Related work}
\label{sec:related}

Ontologies have been used for various applications in the field of automated driving.
A major part of the contributions comes from work in recent years on situation assessment and behavior planning.

Armand et al. \cite{armand_ontology-based_2014} describe how ontologies can be applied to model interactions based on spatio-temporal relations between traffic participants and infrastructure.
They are applied to infer how classified object groups will behave and interact with the host vehicle in the future.
Sensor data are used as assertional boxes for an ontology and to develop a human-like scene understanding.
The understanding is based on object tracks, map data, and the dynamic state of the host vehicle.
Therefore, the concept classes are divided into mobile entities, static entities, and context parameters which describe spatio-temporal relations.
Relations between the classes are described by object properties for actual behavior like \emph{goes towards} or necessary behavior like \emph{has to decelerate} to maintain safe traffic.
To integrate readings from vehicle sensors, the classes are annotated with data properties to map physical measures to the knowledge represented in the ontology.
For the assessment and prediction step, behavior rules are stored in the \emph{semantic web rule language} to infer knowledge from the terminological boxes to a given assertional box from the sensor data. 
The results are taken from an experimental car setup and show that ontology based knowledge can be used to assess traffic scenes in real time applications.

Similar scene understanding approaches are described by earlier contributions in H\"ulsen et al. \cite{hulsen_traffic_2011} and Hummel et al. \cite{hummel_scene_2008}, as well as in Zhao et al. \cite{zhao_ontology-based_2015} \cite{zhao_fast_2016} and Mohammad~et~al.~\cite{mohammad_ontology-based_2015}.

Ulbrich et al. \cite{ulbrich_graph-based_2014} propose an environmental model derived from a knowledge base with hierarchical classes and relations between the entities.
The ontology is implemented in C++ to provide an environmental model, which was updated by sensor data and used to make online decisions in the test vehicle \emph{Leonie} of the project Stadtpilot \cite{nothdurft_stadtpilot}.

%provine et al.\cite{provine_ontology-based_2004}
%\begin{itemize}
%	\item pathplanning with knowledge base 
%	\item vehicle classes and traffic infrastructure
%	\item relations between entities and resulting damage
%	\item influences on damage 
%\end{itemize}

%Barrachina et al. \cite{barrachina_veacon:_2012} uses an ontology in combination with vehicular communication networks to improve traffic safety. 
%The concept of the ontology is designed to represent multiple causes for accidents extracted from the \emph{General Estimates System} of the National Highway Traffic %Safety Administration of the USA (NHTSA). 
%With the knowledge from this ontology the authors are able to automatically determine which vehicle types (trucks, buses, trailer, cars) are endangered by %environmental factors. 
%They extend the knowledge to specific target groups (like age, sex, weight). 
%This knowledge is then used in a simulation to investigate how warning messages for endangered drivers in the specified driving scenarios can improve traffic safety.
%\cite{fuchs_integration_2008}

%\cite{kannan_intelligent_2010}
%\cite{lattner_knowledge-based_2005}
%Pollard et al. \cite{pollard_ontology-based_2013} propose an approach for the classification of automation levels based on several aspects of automated vehicles.
%In the first step the longitudinal and lateral control including abilities for local planning, parking and global planning are assessed.
%Afterwards, the systems are distinguished by their modeled aspects of situation assessment.
%This information is used to infer which automation level a system fulfills.

Geyer et al. \cite{geyer_concept_2014} formulate the need for a unified understanding of terms and definitions for automated driving to generate requirements expressed by use cases or scenarios. 
For this purpose, they propose an ontology to define and order the terms ego vehicle, scenery, scene, situation, scenario, driving mission, and route.
Use cases and scenarios can be defined in a unified representation with the order derived from the ontology.
The ontology is not technically implemented and shall form a basic understanding by reading the publication. 
After proposing the ontology, the authors emphasize two challenges in the generation of requirement catalogs: top-down development and the form of representation. 
On the one hand, information must be as complete as possible.
On the other hand, they shall be understandable by humans involved in the development process. 
The main utilization of the proposed nomenclature and concept is to homogenize two projects in the field of cooperative vehicle guidance. 
A top-down approach for generating the catalogs was not described in the contribution.
However, usage of guidelines for road design was mentioned as a starting point.

The ontology suggested by Geyer et al. was reviewed, unified with other contributions and further defined by Ulbrich et al. \cite{ulbrich_scene_2015}.
They defined differences and the coherence between the properties of each term.
This contribution follows the definitions of Ulbrich et al. for the terms \emph{scene, situation} and \emph{scenario}.

Xiong \cite{xiong_creating_2013} proposes a framework for \emph{scenario orchestration with autonomous simulated vehicles} for simulation of test cases based on driving scenarios for automated vehicles. 
Therefore, the framework consists of an ontology for scenario orchestration (OSO), virtual driver(s), a collection of supporting modules in a scenario management module (SMM), and a scenario observer. 
One of the main concepts is a simulation supervision linked with the driver models, which calculate all interactions between the entities in the simulation and the necessary tools according to the defined scenario depending on the simulation framework.
This framework has been evaluated in two simulation platforms by executing multiple predefined scenarios.

The ontology used in the framework describes concepts and relations between driving context, task representation, (simulative) actions, simulation monitoring, and temporal representations between entities in the simulation.
%\begin{figure}[h]
%	\begin{center}
%	\includegraphics[width=0.45\textwidth]{bilder/xiong}    % The printed column width is 8.4 cm.
%	\caption{Ontology for scenario orchestration from \cite{xiong_creating_2013}} 
%	\label{fig:onto_xiong}
%	\end{center}
%\end{figure}
%Fig. \ref{fig:onto_xiong} shows the main concept for the ontology.
The scenario representation around a vehicle (in this case) consists of road segments, intersections, and vehicles.
Further, the orchestration framework is further able to determine which actions of vehicles are referred to which monitoring devices and measurements.
This allows the framework to represent all of its tasks in relation to the actual simulated scenario.
Nevertheless, scenarios for evaluation have been designed by experts and the thesis has the focus on orchestration of all entities regarding simulation using the knowledge-based system.

To summarize the related work, we assume that ontologies provide a suitable framework for various applications in automated driving.
This includes support of automated driving functions by inferring knowledge about traffic scenarios and enhancements in traffic safety by supporting a centralized traffic system.
In regard of our focus on scene creation, we further investigated the concepts of Geyer et al. \cite{geyer_concept_2014} and Xiong \cite{xiong_creating_2013}.
Ontologies from contributions discussing scene understanding are compared regarding the represented entities and hierarchies of classes to provide a basis for our knowledge modeling approach.
In contrast to approaches for scene understanding from sensor data through ontologies, our approach for scene generation from an ontology needs to derive every possible assertional box from the modeled knowledge represented as terminological boxes.
So far, we did not find contributions utilizing ontologies for initial scene creation.

\section{Ontology-based Scene Creation}

As introduced in the beginning, automated vehicles will have to accomplish a large number of operating scenarios during their lifetime.
Before deploying automated vehicles in public traffic, developers have to investigate these scenarios in the development process in order to ensure safe systems.
Go and Carroll \cite{go_blind_2004} describe this approach as scenario-based design paradigm for complex systems.
The main idea of the concept is that multiple views of the same set of operating scenarios of a system by different stakeholders are needed to get well-engineered requirements.
For automated vehicles, scenario-based design can help to analyze the system from multiple points of view.
This includes defining the behavior of the system, human machine interaction, risk analysis, as well as the test process in simulations before going to expensive field tests.

According to Ulbrich et al. \cite{ulbrich_scene_2015} ``a scenario describes the temporal development between several scenes in a sequence of scenes. Every scenario starts with an initial scene.''
This contribution aims at proposing a creation process for (initial) scenes. 
%First usage for the created scenes is a hazard analysis and risk assessment in the project aFAS where scenes are paired with functional failures and then be assesses by experts.

In a previous contribution \cite{bagschik_identification_2016}, we propose a scene creation for a hazard analysis and risk assessment supported by a database at the example of an unmanned protective vehicle in the project aFAS \cite{stolte_towards_2015}.
Since the outcome of initial scenes will be identified by the experts in the project, we focus on the creation of scenes.
A major disadvantage of combining database entries to generate scenes is that many of them were physically not possible or unreasonable.
Databases contain no semantics which leads to unreasonable scenes.
The project aFAS works, compared to other automated vehicles, in a limited use case where a protective vehicle shall operate unmanned and unsupervised on the hard shoulder of motorways.
Our conclusion is that the process based on databases would not scale for wider use cases and more complex\footnote{Related to quantity of participating elements} scenes.
The observation showed that knowledge is implemented implicitly by considering exceptions in the combination step in the database processing.
For example, the knowledge about which operating mode can be executed in which area of the hard shoulder (normal or on- and off-ramps) was not explicitly designed in the database, but considered in the creation process.
When it comes to arguing the safety of a system, assumptions and knowledge should be represented rather explicitly than implicitly to maintain traceability throughout the development process.

\begin{figure}[h]
	\begin{center}
	\includegraphics[width=0.31\textwidth]{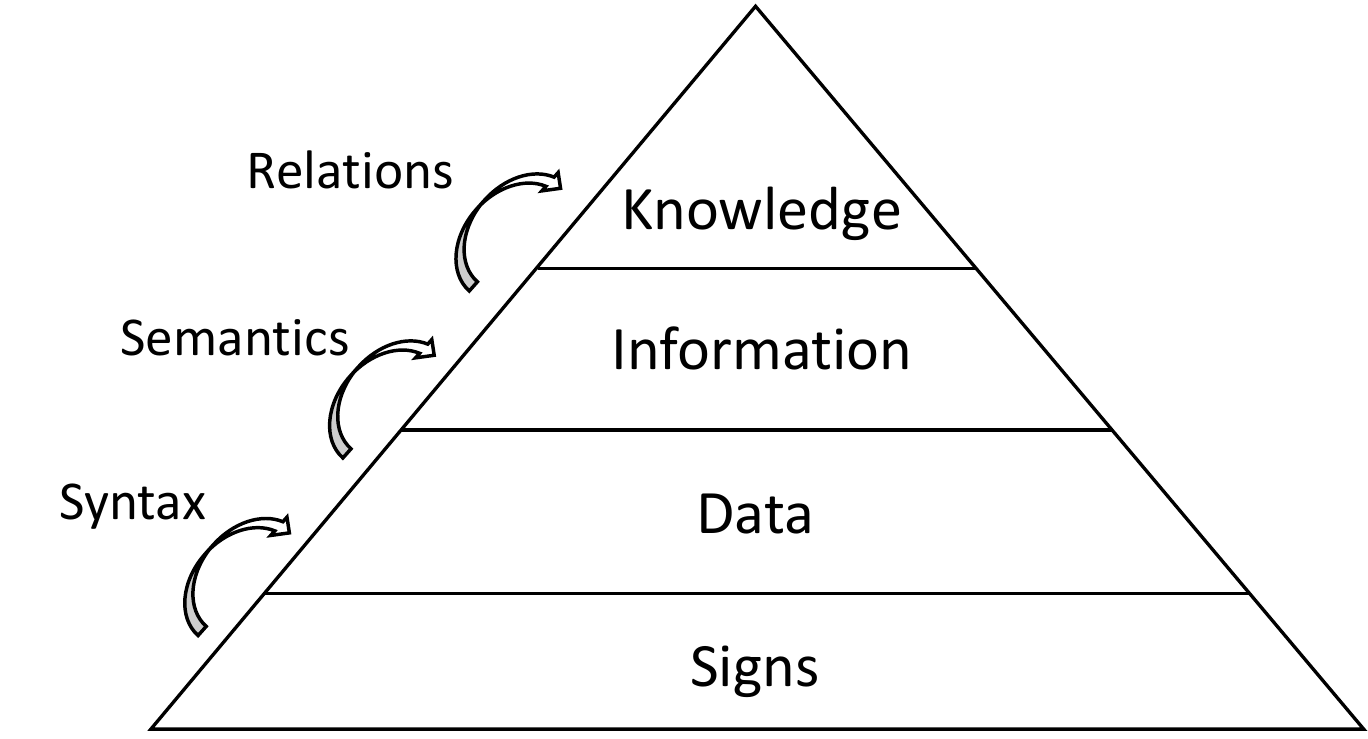}    % The printed column width is 8.4 cm.
	\caption{Hierarchy of signs, data, information and knowledge adopted from Ackoff~\cite{ackoff_data_1989}} 
	\label{fig:wisdom}
	\end{center}
\end{figure}

\begin{figure}[h]
	\begin{center}
	\includegraphics[width=0.40\textwidth]{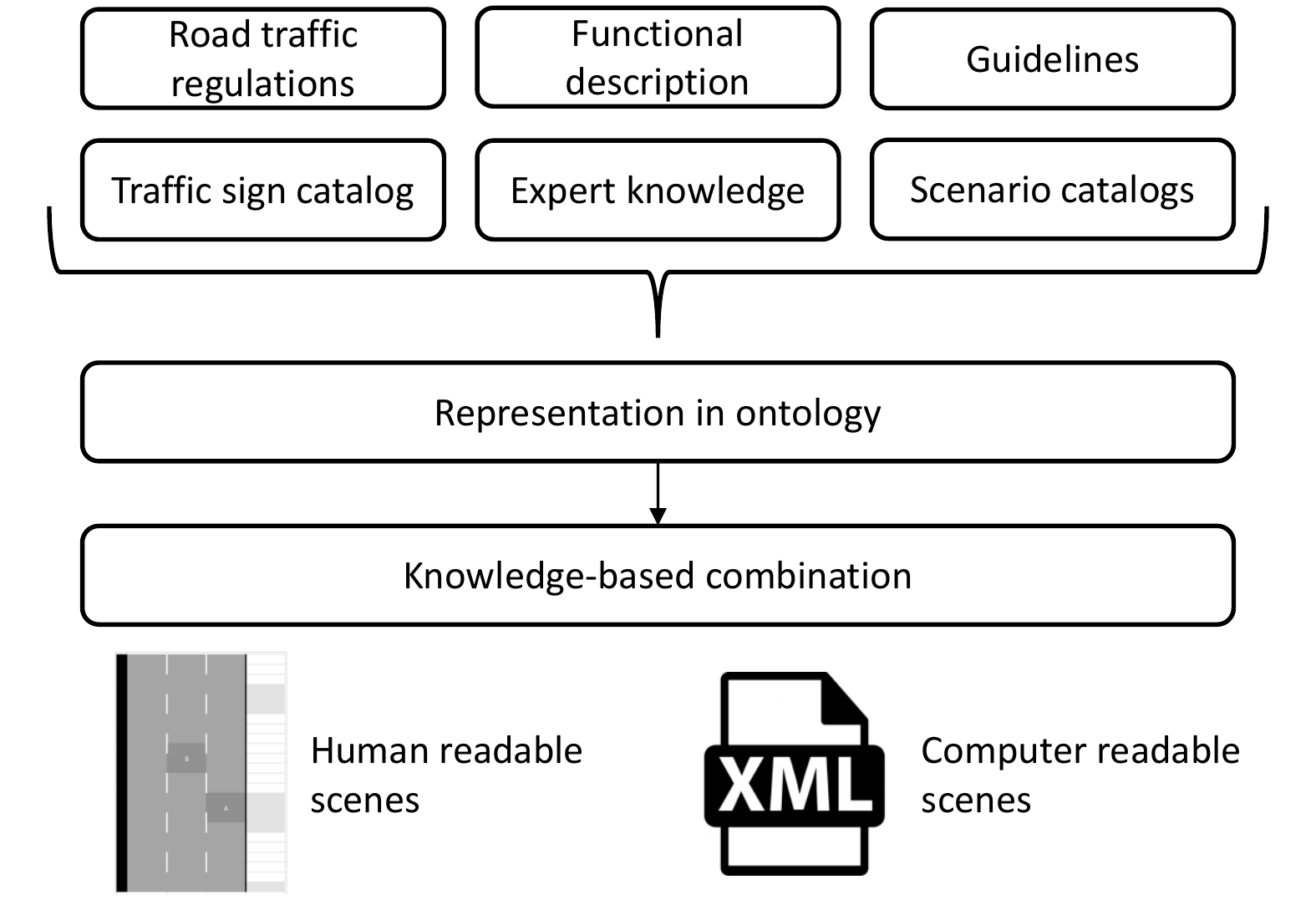}    % The printed column width is 8.4 cm.
	\caption{Ontology-based process for scene creation} 
	\label{fig:process}
	\end{center}
\end{figure}

To get more accurate and useful scenes, we need a system that knows how data can be combined and what meaning data has in the real world.
Ackoff \cite{ackoff_data_1989} describes the hierarchy of data, information, knowledge and wisdom as shown in Fig. \ref{fig:wisdom}.
To conclude the work of Ackoff, instead of data we use information with semantics to combine information to meaningful scenes of real world traffic.
Ontologies provide knowledge bases which are understandable for humans and computers.
Based on the concepts of Ackoff, we propose a process for an ontology based scene creation as shown in Fig. \ref{fig:process}.
The following subsection will explain the knowledge acquisition, how knowledge is modeled, and the combination process.

\subsection{Knowledge acquisition}

Knowledge systems are divided into knowledge acquisition and knowledge representation.
Guidelines for the creation of traffic infrastructure can be used for knowledge acquisition.
These guidelines describe how infrastructure is organized, named, and how the relations can be represented.
In Germany, guidelines for the creation of motorways, rural roads, inner city roads, crash infrastructure, markings, traffic lights, and signs exist and are periodically updated to new regulations.
There are also various scenario catalogs from past projects or system descriptions, which contain knowledge about scenarios.
These sources can also be used if the nomenclature of scenarios is translated properly.
As Geyer et al. \cite{geyer_concept_2014} describe, each project or catalog has its own nomenclature and concepts how to organize knowledge.

Guidelines apply to whole domains of traffic. 
The number of scenes, which can be created by combination of all entities of a domain, is very large.
However, only a selection of scenes depending on the functional range is relevant for investigation.
Therefore, the functional description of the system has influence on the knowledge base. 
When a system is planned to operate automatically on the hard shoulder only, as the vehicle in the project aFAS, scenes in which no hard shoulder is present are not necessary for the evaluation.
However, for the traceability of assumptions made in the analysis, it is important to explicitly represent decisions on which entities or relations from the knowledge are irrelevant for the system. 

The third important aspect besides guidelines for traffic domains and the functional description is expert knowledge.
On the one hand, every guideline defines only a part of traffic infrastructure and we need to represent how these parts can interact and which dependencies they have to each other.
For example traffic rules regarding individual speed limits for each lane on motorways are bound to at least three lanes for each driving direction.
On the other hand, we need to represent real world deviations from the guidelines and very unlikely entities or combinations of these.
Reschka \cite{reschka_safety_2016} describes safety-relevant events and how dilemma scenarios can occur from event chains in traffic.
Knowledge about the causes of event chains is important to identify functional boundaries of the item and to derive meaningful safety concepts.

\subsection{Layered model for knowledge representation}

\begin{figure}[h]
	\begin{center}
	\includegraphics[width=0.4\textwidth]{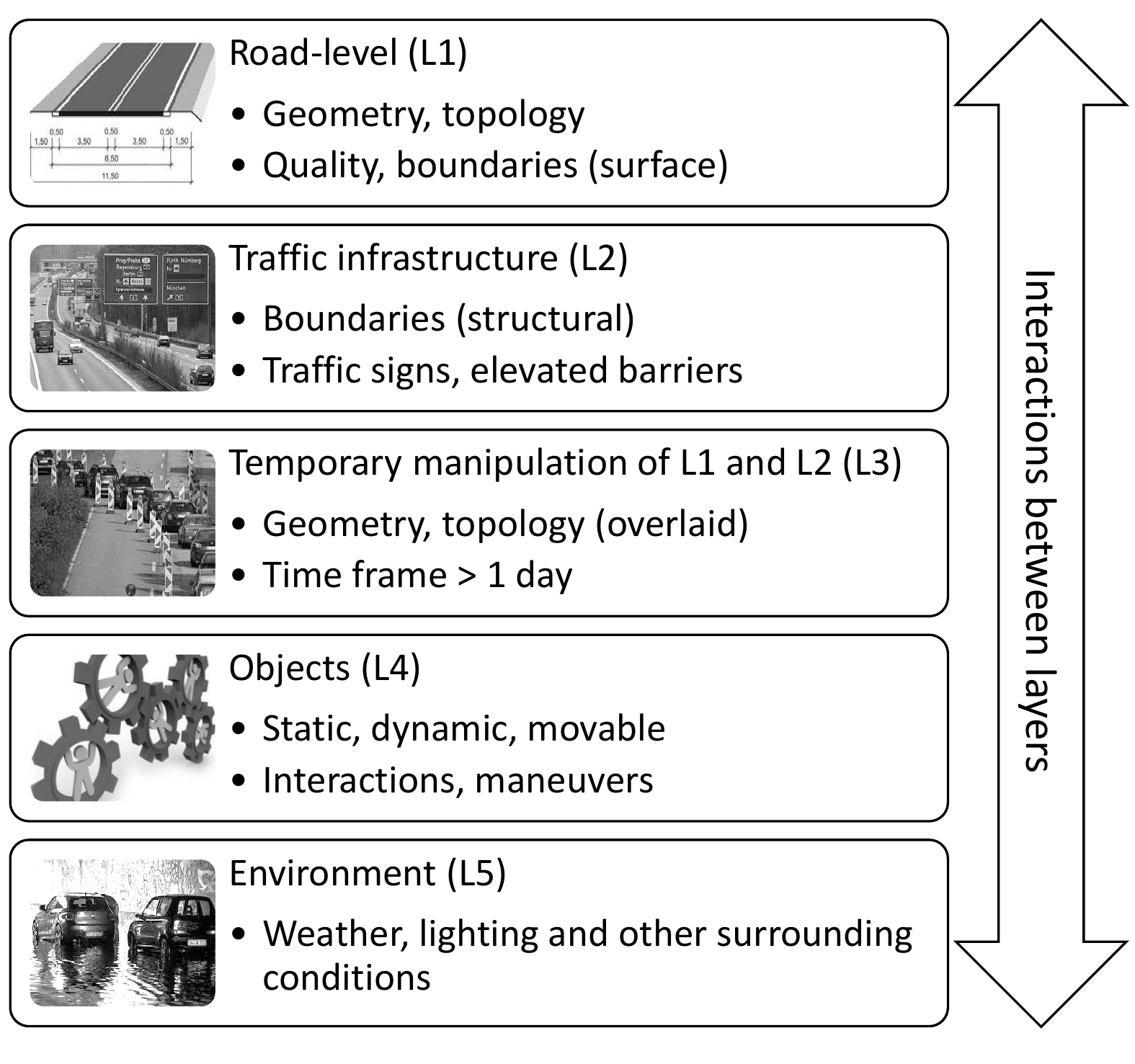}    % The printed column width is 8.4 cm.
	\caption{Layer model for the representation of driving scenes adopted from Schuldt \cite{schuldt_beitrag_2017}} 
	\label{fig:layers}
	\end{center}
\end{figure}

To organize all information in a knowledge base, we propose to use an adapted layered model for scene representation based on the work of Schuldt \cite{schuldt_beitrag_2017} as shown in Fig. \ref{fig:layers}.
The first layer describes the layout of the road, including markings and topology. 
Schuldt \cite{schuldt_beitrag_2017} proposes to use basic elements like straights, curves and clothoides to define geometries.
The parameters for the geometric models can be derived from guidelines in each domain of traffic (motorways, rural roads and urban roads).
Further, the basic layer  defines how the geometries are related to each other to build up the topology.
For example, motorways in Germany follow defined layouts where markings, parameters and relations are given by standards.
These guidelines can be used to model hierarchical concepts on how geometries representing lanes can be put together to build valid road snippets for traffic scenes.

The second layer adds traffic infrastructure to the road-level.
Traffic infrastructure is part of the scenery, but not all stationary objects belong to the traffic infrastructure.
This differs from the four-layered concept of Schuldt \cite{schuldt_beitrag_2017} as we focus on separating semantic relations for an automated creation of scenes.
For the creation process we propose to model traffic rules like speed limits or no passing zones as top-level classes.
The sub-classes then represent instantiations of the traffic rules through traffic infrastructure like signs or road markings, which express the rules for the driver. 
Instantiations of traffic rules have several dependencies on each other which can be extracted from road traffic regulations.
Due to dependencies on markings, the influences on the first layer have to be modeled according to the traffic rules.

Temporary manipulations of the first two layers are represented in the third layer.
%We propose to distinguish between manipulations and placed objects by the time the manipulation persists.
In this concept, the time frame for temporal manipulations was chosen to one day.
Other objects which persist longer than one day can be modeled as infrastructure on the layers below.
The layer contains classes which represent how construction sites have to be marked, routed and secured.
The resulting changes to the original layout are marked as manipulation in the resulting traffic scenes.

All objects which do not necessarily belong to the traffic infrastructure are modeled in the fourth layer.
Stationary and movable objects can be placed without extensive changes in the relations of the traffic infrastructure.
Traffic participants are categorized into object classes like cars, trucks, cyclists, etc.
To define interactions between the participants, we use atomic maneuvers which are disjoint to each other.
Based on T\"olle \cite{tolleFahrmanoverkonzeptFurMaschinellen1996}, Reschka \cite{Reschka2016} proposes to separate the whole driving task into nine maneuvers:
drive up, follow, approach, pass, lane change, turn, turn back, and safe stop.

Each of these maneuvers is described by relational parameters to surrounding objects and semantic rules.
For example the maneuver \emph{approach} can be described by the distance to the approaching object, relative speed and derived parameters like time-to-collision or time-headway.
If the rear vehicle of two following vehicles is driving fast in the same lane this maneuver can be categorized as approaching.
These rules can be expressed by the semantic web rule language and extend the conceptualization of classes of traffic participants by behavior rules.
Maneuvers can also have relations to other layers if a car is approaching some infrastructure.
For our approach we extended the maneuvers with \emph{fall back} as opposite to approach to describe a constellation where a vehicle behind another one has a lower speed.

In the fifth layer, Schuldt \cite{schuldt_beitrag_2017} describes environmental effects like weather but also the influences on infrastructure like aquaplaning.
In our concepts, environmental effects will also consider influence on the interactions between traffic participants which results in different parameter ranges for relative parameters.

\begin{figure}[h]
	\begin{center}
	\includegraphics[width=0.37\textwidth]{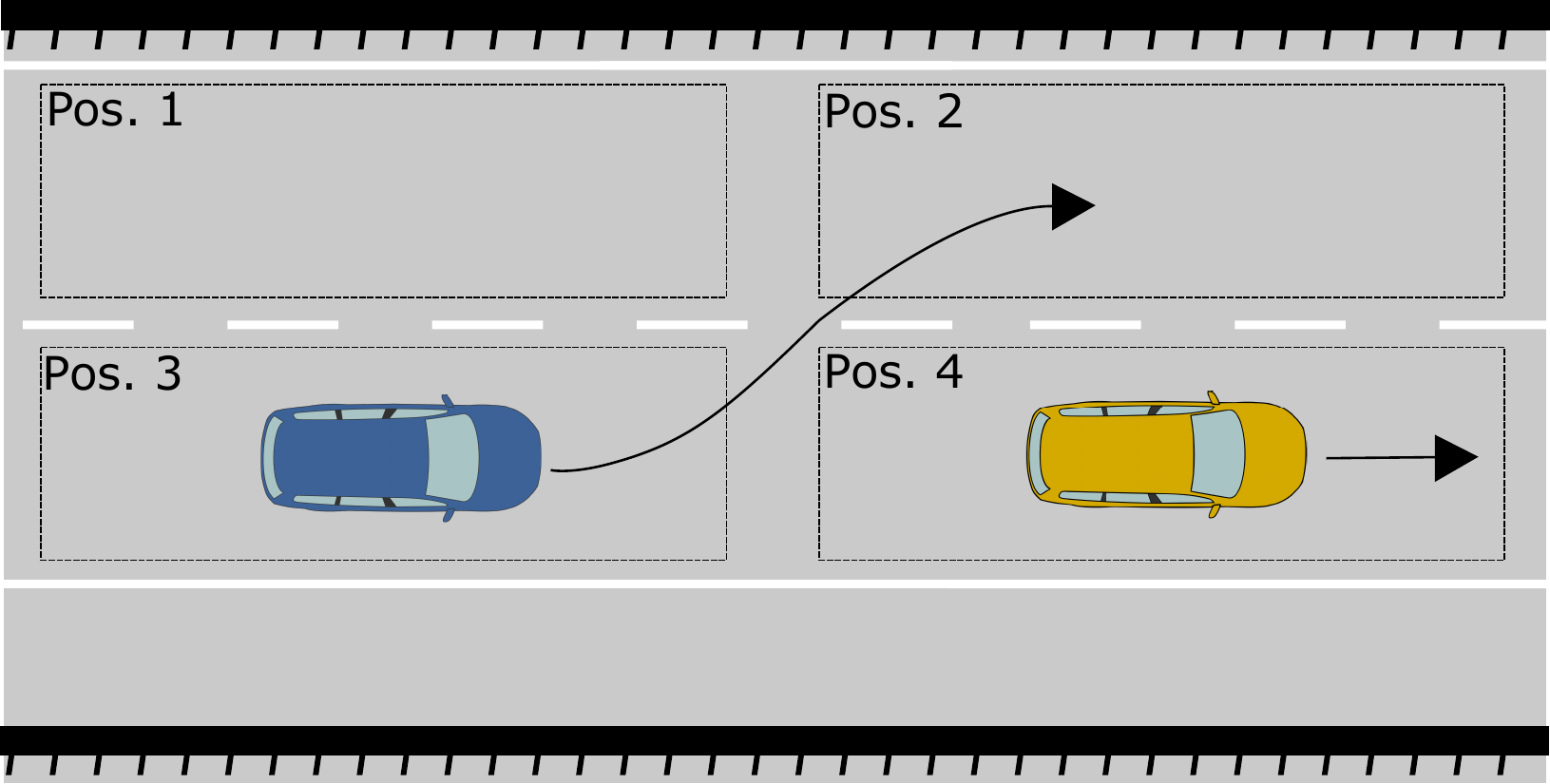}    % The printed column width is 8.4 cm.
	\caption{Example scene} 
	\label{fig:ex_scene}
	\end{center}
\end{figure}

Fig.~\ref{fig:ex_scene} shows a simple scene including:
\begin{itemize}
	\item Layer 1: Layout RQ 31 with 2 lanes and a hard shoulder \cite{RAA}
	\item Layer 2: Solid lines on the track limits, dashed line between lanes and solid crash barriers besides the road
	\item Layer 3: Nothing
	\item Layer 4: Blue vehicle starting a lane change to the left behind a yellow car which follows the right lane
	\item Layer 5: Normal weather and temperature conditions (not visible in the sketch)
\end{itemize}

\subsection{Process for creation of traffic scenes}

After the ontology for the given use case is modeled, the process for creation uses the knowledge to create valid traffic scenes.
The ontology clusters entities in natural language and assigns a formal order through conceptualization.
We propose to model parameter relations with data properties in the ontology.
Each entity (represented by a word) in the ontology represents multiple relations to parameters in the physical state space.
Lanes in the traffic network may include parameters like width, condition and friction coefficient.
To represent interactions of layers, we annotate if an entity includes or influences a parameter.
For example, rain in the weather layer has influence on the friction coefficient, which is included in the lanes in the first layer.

For the creation of all possible traffic scenes from a given ontology, we need to derive all possible assertional boxes.
This stays in contrast to the contributions presented in Section \ref{sec:related}.
All approaches aim at inferring knowledge from terminological boxes to augment observed assertional boxes.

\begin{figure}[h]
	\begin{center}
	\includegraphics[width=0.35\textwidth]{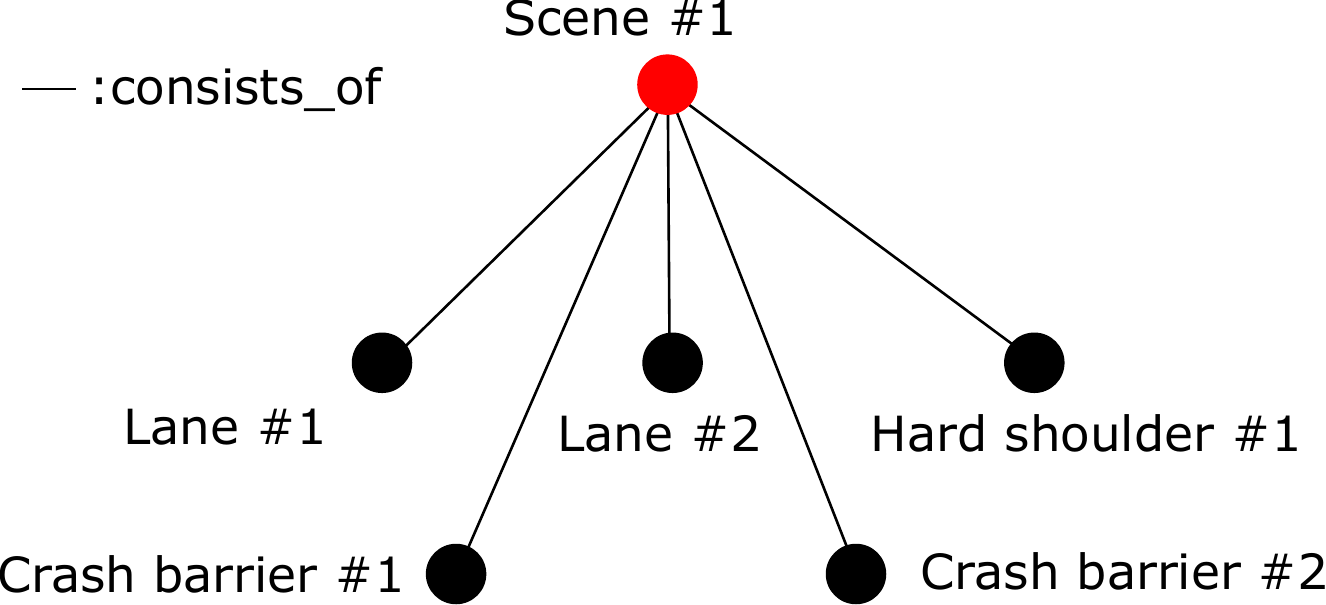}    % The printed column width is 8.4 cm.
	\caption{First step of scene creation containing existential information} 
	\label{fig:ex_layer1}
	\end{center}
\end{figure}

The traffic scene creation starts with layer 1 and 2 from the layered model. 
All concepts of possible road layouts according to German guidelines are stored in our ontology.
Each layout is described by mandatory and optional elements.
The RQ 31 layout from Fig.~\ref{fig:ex_scene} at least consists of two lanes and a hard shoulder, and optional barrier and embankment elements as well as possible traffic rules like speed limitations.
Our ontology uses statements like \emph{consists\_of}, \emph{has\_optional}, \emph{enables}, and exact cardinalities for modeling possible road network layouts.
For the automated creation process we use the java-based OWL-API\footnote{http://owlcs.github.io/owlapi/} to access the elements and relations and build up every valid combination allowed by the modeled rules.
Each layout then is stored in a separate assertional box with concrete instances where traffic scenes can take place.
The layout from our example in Fig.~\ref{fig:ex_scene} would then be expressed by Fig~\ref{fig:ex_layer1}.

\begin{figure}[h]
	\begin{center}
	\includegraphics[width=0.35\textwidth]{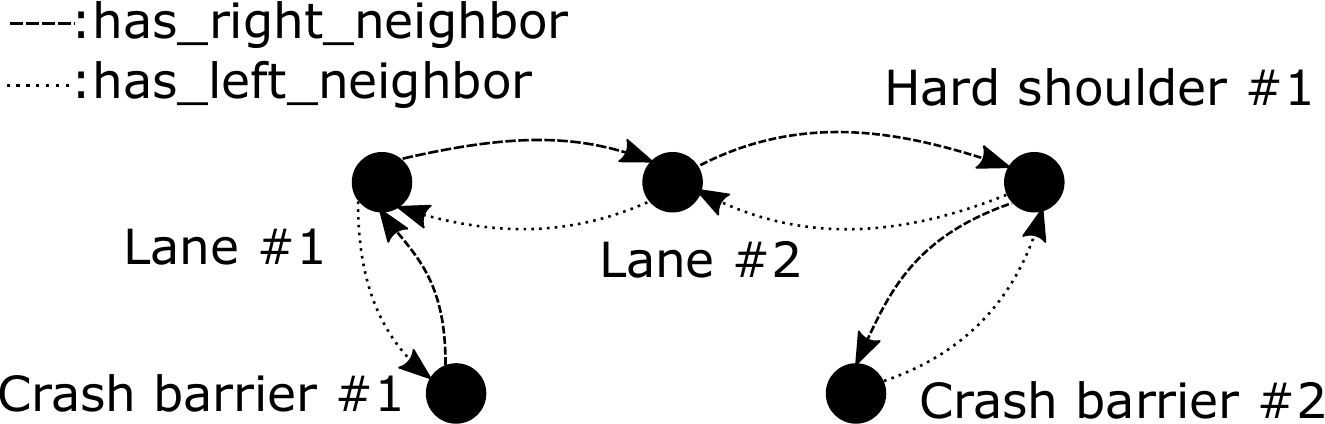}    % The printed column width is 8.4 cm.
	\caption{Second step of scene creation containing arrangement information} 
	\label{fig:ex_layer1_arra}
	\end{center}
\end{figure}

The generated infrastructure scenes only contain information which elements exists in a certain scene.
To gain a useful scene description, the instances have to be arranged. 
This is the first step of logic reasoning in our process.
Elements can be arranged \emph{right}, \emph{left}, \emph{in\_front\_of}, and \emph{behind} from each other.
Thereby right and left as well as in\_front\_of and behind are reflexive properties.
That means, that if one element is right of another, the second element automatically is left of the first one.
Arrangement properties are only inferred for direct neighbors in the scene, which results in the properties shown in Fig.~\ref{fig:ex_layer1_arra} for our example.

The third layer of the layered model is not implemented in the first evaluation of our approach.

\begin{figure}[h]
	\begin{center}
	\includegraphics[width=0.32\textwidth]{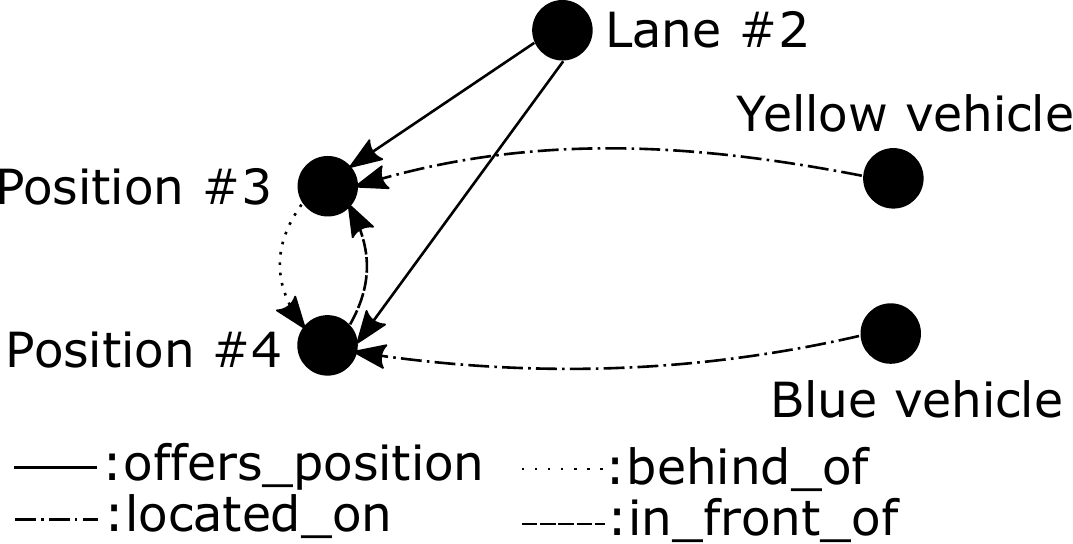}    % The printed column width is 8.4 cm.
	\caption{Third step of scene creation containing traffic participants and arrangement information} 
	\label{fig:ex_layer1_cars}
	\end{center}
\end{figure}

For the fourth layer we use a combination of permutation and logic reasoning to generate traffic participants and the possible behavior in regard of the given infrastructure and traffic rules.
Therefore, a configurable number of positions per lane is distributed on the infrastructure. 
Fig.~\ref{fig:ex_scene} shows two possible positions on each lane of the scenery.
The positions are generated based on a relation (\emph{offers\_position}) of the lane classes of the ontology.
Each lane then gets the defined number of instances of positions (number 1 to 4 in our case).
The arrangement of the position instances is inferred similar to ones before and based on the arrangements of the infrastructure.
So far, only lanes offer positions, but for future adaptations of the approach the hard shoulder, for example, could offer positions for standing vehicles too.
The second step then adds a defined number of traffic participants on the positions and permutes them to possible combinations.
Fig.~\ref{fig:ex_layer1_cars} shows the inferred scene for the example from Fig.~\ref{fig:ex_scene}.

This step considers elimination of reputations for semantic classes. 
In our example, it would make no difference if the blue vehicle is behind the yellow one or the other way around (semantic reputation).
If the example would contain a truck and vehicle (passenger car) both combinations would make up another scene and would have to be generated.
After the traffic participants are located relative to each other, driving maneuvers can be inferred from the positioning. 
For each maneuver a semantic web rule has to be implemented in the ontology. 
Semantic web rules can combine logic operators into rules.
For example, a lane change to the left is possible if the car which shall conduct the maneuver has a neighboring lane on the left and has no car on the positions beside and in front of it in the neighboring lane.
Based on the arrangement, every possible maneuver (follow, approach, fall back, lane change right/left, start from stand) for every traffic participant in the scene is inferred.
In the last step, invalid or forbidden (expressed by semantic web rules) combinations are eliminated from the scene catalog.
Forbidden maneuvers are based on the traffic rules from layer 2 (for example forbidden overtaking for trucks) and invalid combinations are given by the creation process. 
If potentially critical scenes shall be assessed, a scene in which two cars can change the lane towards the same position on the infrastructure could be generated. 
For normal (comfort) driving scenes, this combination would be invalid.

In the fifth layer predefined weather setups (sunny, rainy, cloud, etc.) are modeled and permuted at the end of the creation.
For future investigations weather conditions will have recursive influence on allowed maneuvers and traffic rules (e.g. speed limits during wet road surface).

\section{Preliminary Results and Discussion}

\begin{figure}[h]
	\begin{center}
		\includegraphics[width=0.9\columnwidth]{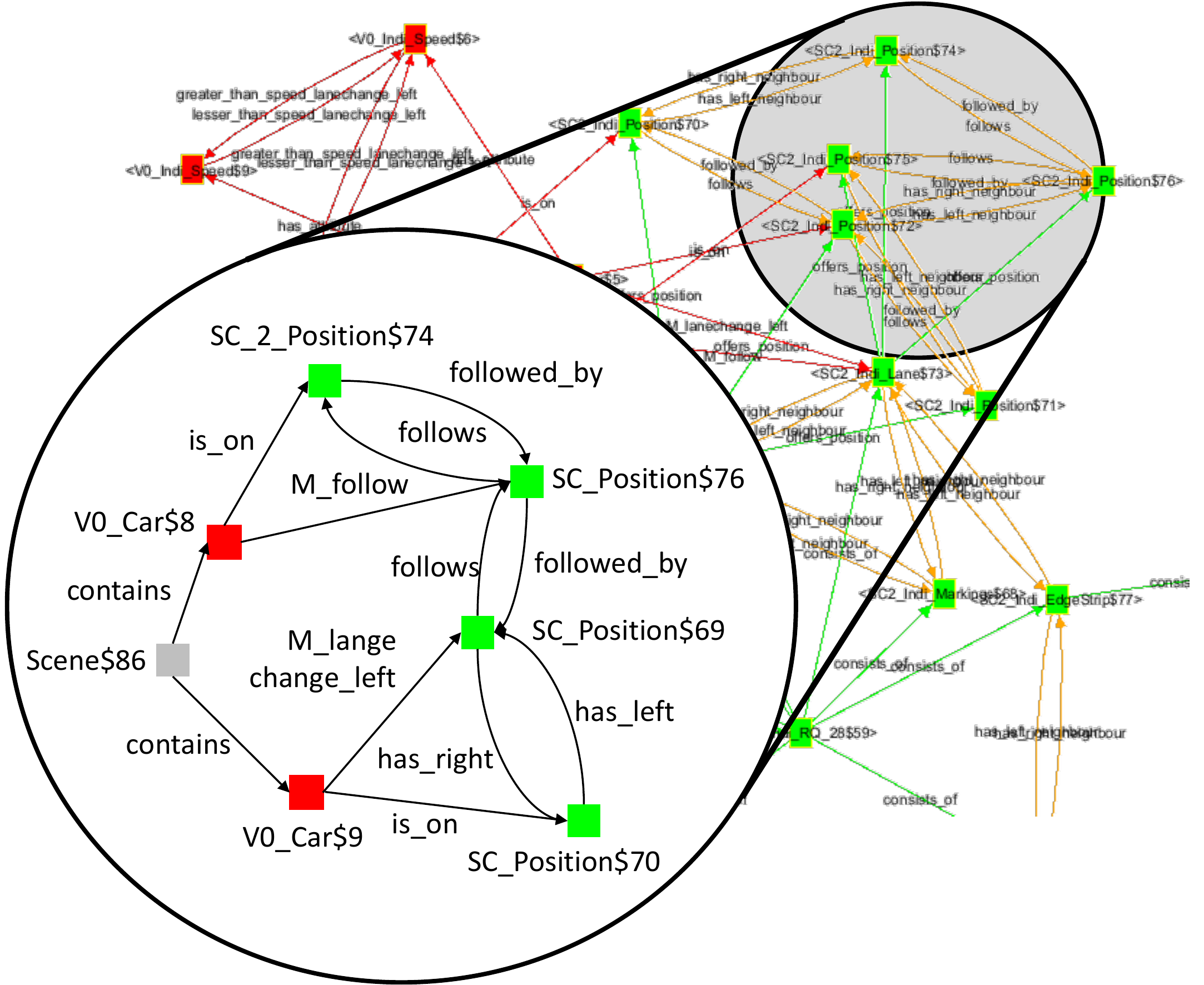}    % The printed column width is 8.4 cm.
		\caption{Example scene graph. Red squares mark vehicles, green squares mark positions on the infrastructure, gray square marks the scene, arrows and captions describes relations and maneuvers.} 
		\label{fig:example_graph}
	\end{center}
\end{figure}

To evaluate the concept, we modeled an ontology with 284 classes, 762 logical axioms and 75 semantic web rules to generate scenes for German motorways.
Depending on the number of positions per lane in the infrastructure and the number of traffic participants we are able to automatically generate a large number of scenes. 
These created scenes in natural language can be used to investigate and define safe behavior of an automated vehicle in the concept phase or to derive test cases for simulation environments with a high range of varieties. 
Fig.~\ref{fig:example_graph} shows a part of a scene from the automated creation process.

\begin{figure}[h]
	\begin{center}
		\includegraphics[width=0.9\columnwidth]{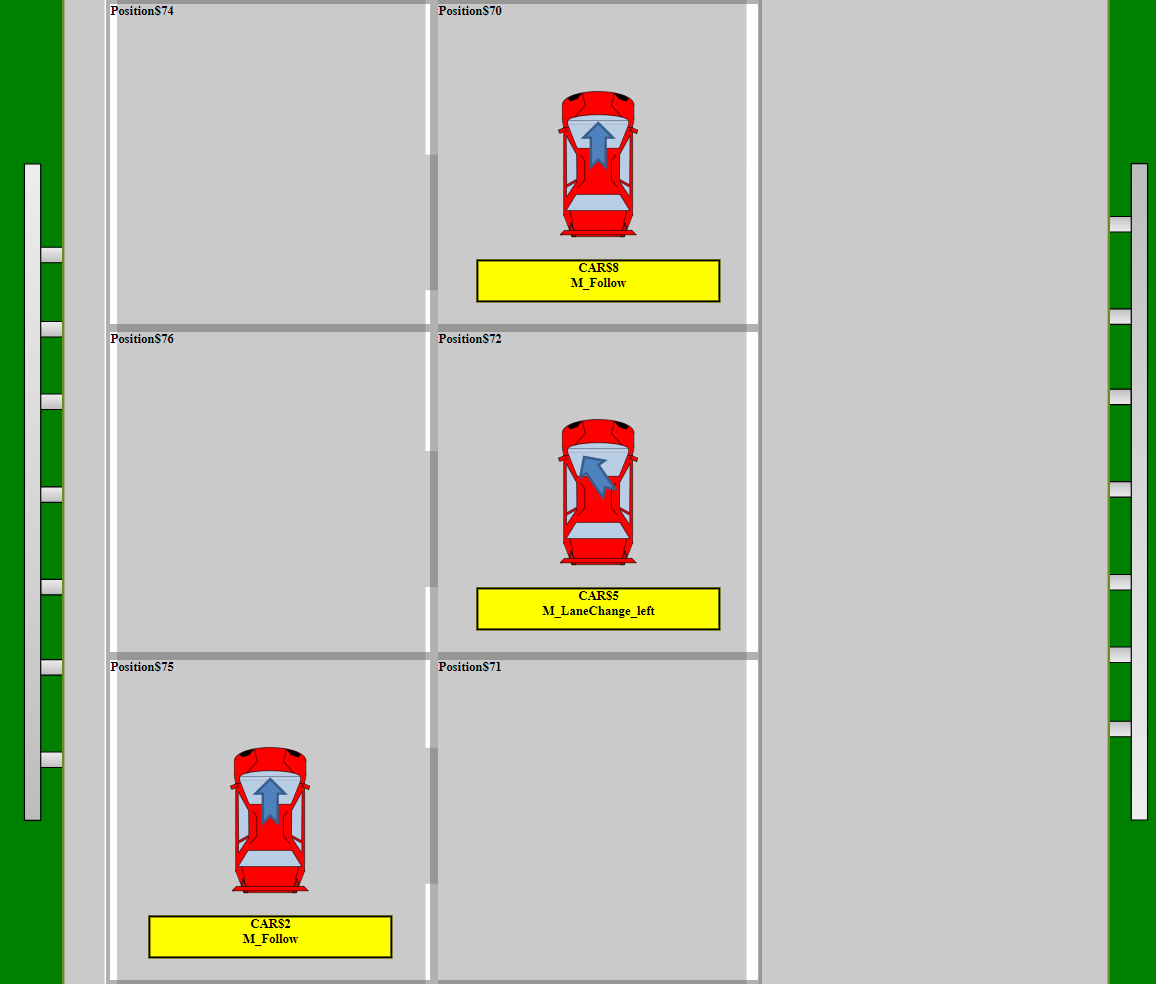}    % The printed column width is 8.4 cm.
		\caption{Example HTML bases visualization of an automatically generated scene.} 
		\label{fig:html}
	\end{center}
\end{figure}

Red squares mark vehicles, green squares mark positions on the infrastructure, gray square marks the scene, arrows and captions describes relations and maneuvers.
For a first evaluation of this contribution we created a typical infrastructure setup on a three-lane motorway with three vehicles and positions per lane in good weather conditions.
Therefore, we got 1016 traffic scenes which contain all combinations of three vehicles with possible maneuver combinations on the motorway.
As introduced, we successfully removed semantic duplications, as the vehicle are permuted with repetition in the same semantic classes.
Without this adjustment the number of possible scenes would be more than three times larger.
Furthermore, we generated different combinations of Layer~1 and 2 for two-, three-, and four-lane motorways with different traffic rules.
For these examples we got a total number of 652 infrastructure setups without combinations of traffic participants.
These number indicate how large a full combination of all layers would be if three or more traffic participants are combined with 652 infrastructure setups.
Since the complete description of scenario in natural language is still a large document, we also developed a simple HTML based visualization as shown in Fig.~\ref{fig:html}.
This gives an user a quick overview of the infrastructure and the vehicles with maneuvers in the scene.

As a major advantage we see the reduced complexity in the knowledge base compared to the resulting scene catalog.
To find missing elements on the layers, analysts do not need to investigate the scene catalog.
Instead, it is possible to quickly access knowledge on the layers and check the modeled constraints and rules.
If the elements are correctly modeled in the ontology, the elements will also be correctly combined with other layers in the resulting scene catalog.

Besides all combinatorial creation of traffic scenes, the approach lacks some disadvantages against real world scenarios.
For now, the positioning in the lanes do not take a vehicles dimensions into account.
This excludes, for example, two motorcycles riding next to a truck, which has multiple times a length of a motorcycle.
There are also possible infrastructure setups which do not exactly apply to rules of guidelines.
These cases can only be identified by real world data, which then may lead to new semantic rules and constraints in the knowledge base.
To conclude our results, we propose to use our approach complementary to real world data analysis and as a base for creation of broadly variated traffic scenes for e.g. simulation based testing.

\section{Conclusion and Future work}

In this contribution we proposed an approach for a knowledge based scene creation for automated vehicles.
Ontologies are a representation of knowledge-based systems, which are widely applied in semantic web applications.
Publications from the recent years show that ontologies also can represent traffic scenes.
In contrast to scene understanding approaches for automated vehicles, we proposed a method to derive possible observations (assertional boxes) from modeled knowledge (terminological boxes).

A main advantage over creative expert based scene creation methods is the automated creation based on formalized knowledge.
For the validation and verification, engineers do not have to review the resulting scene catalog but only the modeled knowledge base. 

Based on the fully connected scene graph, we will further investigate the transformation of natural language based scenes to simulation data formats like OpenScenario and OpenDrive.
To make the scene catalog handlebar for human experts, we need to identify valid abstractions concepts to conduct analysis steps on the scene catalog (for example in a hazard analysis and risk assessment).

% use section* for acknowledgment
\section*{Acknowledgment}

%We would like to thank our partners from the project consortium consisting of MAN Truck - Bus AG (consortium leader), TRW Automotive GmbH (ZF TRW), WABCO Development GmbH, Robert Bosch Automotive Steering GmbH, Hochschule Karlsruhe, Hessen Mobil - Road and Traffic Management, and BASt - Federal Highway Research Institute for their support. 
%The project is partially funded by the German Federal Ministry for Economic Affairs and Energy.
We would like to thank the project members of the projects PEGASUS and aFAS funded by the German Federal Ministry for Economic Affairs and Energy for the productive discussions and the feedback on our approaches.
Our work is partially funded by the Volkswagen AG.

We also like to thank Christian K\"orner for his support in implementing and evaluating the concepts.

\IEEEtriggeratref{12}
\bibliographystyle{IEEEtran}
% argument is your BibTeX string definitions and bibliography database(s)
\bibliography{biblio}

% that's all folks
\end{document}